\documentclass{article}

\usepackage[preprint]{neurips_data_2023}

\usepackage[utf8]{inputenc} 
\usepackage[T1]{fontenc}    
\usepackage[export]{adjustbox}
\usepackage{hyperref}       
\usepackage{url}            
\usepackage{booktabs} 
\usepackage{graphicx}
\usepackage{makecell}
\usepackage{microtype}      
\usepackage{multirow}
\usepackage{nicefrac}
\usepackage{subcaption}
\usepackage{xcolor}         

\usepackage{algorithm}
\usepackage{algorithmic}
\usepackage{amsfonts}
\usepackage{amsmath}
\usepackage{amssymb}
\usepackage{amsthm}
\usepackage{bm}
\usepackage{mathtools}
\usepackage{listings}
\usepackage{cleveref}

\title{Hybrid Graph: A Unified Graph Representation with Datasets and Benchmarks for Complex Graphs}

\author{%
    Zehui Li\thanks{Equal contribution. Correspondence to \texttt{\{zehui.li22,x.zhao22\}@imperial.ac.uk}.}\ \ \textsuperscript{1}, 
    Xiangyu Zhao\footnotemark[1]\ \ \textsuperscript{2}, 
    Mingzhu Shen\textsuperscript{2}, 
    Guy-Bart Stan\textsuperscript{1}, 
    Pietro Li\`{o}\textsuperscript{3}, 
    Yiren Zhao\textsuperscript{2} \\
    \textsuperscript{1}Department of Bioengineering, Imperial College London \\
    \textsuperscript{2}Department of Electrical and Electronic Engineering, Imperial College London \\
    \textsuperscript{3}Department of Computer Science and Technology, University of Cambridge \\
}

\begin{document}

\maketitle

\begin{abstract}
Graphs are widely used to encapsulate a variety of data formats, but real-world networks often involve complex node relations beyond only being pairwise. While hypergraphs and hierarchical graphs have been developed and employed to account for the complex node relations, they cannot fully represent these complexities in practice. Additionally, though many Graph Neural Networks (GNNs) have been proposed for representation learning on higher-order graphs, they are usually only evaluated on simple graph datasets. Therefore, there is a need for a unified modelling of higher-order graphs, and a collection of comprehensive datasets with an accessible evaluation framework to fully understand the performance of these algorithms on complex graphs. In this paper, we introduce the concept of \emph{hybrid graphs}, a unified definition for higher-order graphs, and present the \emph{Hybrid Graph Benchmark (HGB)}. HGB contains 23 real-world hybrid graph datasets across various domains such as biology, social media, and e-commerce. Furthermore, we provide an extensible evaluation framework and a supporting codebase to facilitate the training and evaluation of GNNs on HGB. Our empirical study of existing GNNs on HGB reveals various research opportunities and gaps, including (1) evaluating the actual performance improvement of hypergraph GNNs over simple graph GNNs; (2) comparing the impact of different sampling strategies on hybrid graph learning methods; and (3) exploring ways to integrate simple graph and hypergraph information. We make our source code and full datasets publicly available at \url{https://zehui127.github.io/hybrid-graph-benchmark/}.
\end{abstract}


\setcounter{footnote}{0}
\section{Introduction}

Graphs are powerful tools for capturing relationships between objects, ranging from social networks,  biology~\citep{Rozemberczki2021MUSAE} to e-commerce~\citep{mcauley2015image,he2016ups,ni2019justifying}. However, real-world large-scale networks often involve complex node relations beyond only being pairwise. While hypergraphs and hierarchical graphs~\citep{McCallum2000Cora,Getoor2005CiteSeer,Sen2008PubMed,Chien2022YouAreAllSet} have been introduced to capture these complex node relations, they still cannot fully represent the real-world scenarios. Although hypergraphs can describe clustering relations between multiple nodes, they are still limited in their ability to represent hierarchical relations. On the other hand, while hierarchical graphs can make up for the limitation of hypergraphs, they are not flexible enough to represent multi-node relations.

Existing Graph Neural Networks (GNNs) have been proposed for representation learning on simple graphs~\citep{kipf2016semi,hamilton2017inductive,velivckovic2017graph,li2019deepgcns,brody2021attentive} and higher-order graphs~\citep{feng2019hypergraph,yadati2019hypergcn,chan2020generalizing,dong2020hnhn,zeng2019graphsaint,Bai2021HyperConvAtten,brody2021attentive}. 
However, they are usually only evaluated on simple graph datasets, which do not fully reflect the complexity of real-world graph structures. Besides, without a unified modeling of higher-order graphs, nor a collection of comprehensive datasets with an accessible evaluation framework, we cannot fully uncover the underlying performance of these GNNs on complex graphs.

To address these limitations, we endeavour to construct the \emph{Hybrid Graph Benchmark (HGB)}, with a unified view of complex graphs including both datasets and an evaluation framework. Firstly, we introduce the concept of \emph{hybrid graphs}, which provides a unified definition for higher-order graphs. Hybrid graphs extend the notion of simple graphs by allowing nodes to be connected to sets of other nodes, rather than just individual nodes. This enables hybrid graphs to capture more complex node interactions, including simple and higher-order interactions. Based on this unified definition, we build 23 real-world hybrid graph datasets across various domains such as biology, social media, and e-commerce. 
Compared with other existing hypergraph datasets \citep{McCallum2000Cora,Getoor2005CiteSeer,Sen2008PubMed,Chien2022YouAreAllSet} that are on a relatively small scale and do not facilitate hierarchical node relations, we believe our HGB represents a significant step forward in the development of comprehensive datasets for evaluating higher-order graph learning algorithms.

Furthermore, to facilitate fair evaluation of GNNs on our proposed hybrid graph datasets, we provide an extensible evaluation framework and a supporting codebase. Our evaluation framework includes several common graph prediction tasks using their corresponding evaluation metrics, such as node classification and regression tasks. We benchmark seven widely-used GNN models~\citep{kipf2016semi,hamilton2017inductive,velivckovic2017graph,brody2021attentive,zeng2019graphsaint,Bai2021HyperConvAtten,brody2021attentive} on HGB, and introduce three novel baselines that incorporate simple and higher-order graph information, thereby enabling the researchers to conveniently evaluate their own models and compare the results. 
In addition, our empirical study of existing GNNs on HGB reveals various research opportunities and gaps, which are elaborated in later sections. 

To summarise, we provide a comprehensive and unified framework for modelling and evaluating hybrid graph methods, in the hope of stimulating further research in this field. The source code and complete datasets of HGB are publicly available. Our main contributions in this paper are as follows:

\begin{itemize}
    \item We introduce the \emph{hybrid graphs}, a unified view of high-order graphs for fostering further study in representation learning on complex graphs. 
    \item Inspired by this unified mathematical framework, we construct the \emph{Hybrid Graph Benchmark (HGB)} consisting of 23 datasets covering a wide range of real-world applications. We then extend HGB with an easy-to-use and extensible evaluation framework.
    \item Through extensive experimentation, we have verified both the necessity and superiority of our proposed datasets and benchmarking tool. We also draw insights for the graph representation learning community, such as (1) existing hypergraph GNNs may not keep outperforming simple graph GNNs on large-scale networks; (2) appropriate sampling strategies improve the performance of GNNs on higher-order graphs; and (3) integrating simple and higher-order graph information can significantly enhance the prediction performance on complex graphs.
\end{itemize}
\section{Related Work} 

\paragraph{Graph Neural Networks on Simple Graphs} Graph Neural Networks (GNNs) on simple graphs encode the nodes through neural networks, and learn the representations of the nodes through message-passing within the graph structure. GCNs~\citep{kipf2016semi} incorporate the convolution operation into GNNs. 
GAT~\citep{velivckovic2017graph} and GATv2~\citep{brody2021attentive} are another family of GNN variants that improves the expressive power of GNNs through attention mechanisms. GraphSAGE~\citep{hamilton2017inductive} is a general inductive framework that leverages node information to efficiently generate node embeddings for previously unseen data. GraphSAINT~\citep{zeng2019graphsaint} emphasises the importance of graph sampling-based inductive learning method to improve training efficiency, especially for large graphs. While these models succeed in simple graph datasets, it is also of great research interest to test their performance on higher-order graphs. However, there lacks a systematic evaluation of these models on higher-order graph datasets.

\pagebreak 
\paragraph{Graph Neural Networks for Higher-Order Graphs} Higher-order graphs are designed to capture more complex node relations, with hypergraphs being the dominant practice in deep graph representation learning. Hypergraphs refer to graphs in which an edge can connect two or more nodes. In general, GNNs for hypergraphs optimise the node representation through a two-step process. Initially, the node embeddings within each hyperedge are aggregated to form a hidden embedding of each hyperedge. Subsequently, the hidden embeddings of hyperedges with common nodes are aggregated to update the representations of their common nodes. Both HGNN~\citep{feng2019hypergraph} and HyperConv~\citep{Bai2021HyperConvAtten} precisely follow this process. The expressiveness of hypergraph GNNs could be enhanced by modifying this procedure. For instance, HyperGCN~\citep{yadati2019hypergcn} refines the node aggregation within hyperedges using mediators~\citep{chan2020generalizing}; HyperAtten~\citep{Bai2021HyperConvAtten} proposes to use real value to measure the degree to which a vertex belongs to a hyperedge; HNHN~\citep{dong2020hnhn} applies nonlinear functions to both node and edge aggregation processes. 

Hierarchical graphs refer to graphs where nodes are organised into multiple levels, and exhibit hierarchical relations. Compared to hypergraphs, hierarchical graphs are less explored in higher-order graph representation learning. To integrate node information across different levels in the hierarchy, HACT-Net \citep{Pati2022Hierarchical} treats nodes in different levels as separate subgraphs. In HACT-Net, the input features of the lower-level nodes are first passed through a GNN to produce the lower-level node embeddings. Then, the computed lower-level node embeddings are concatenated with the input features of the higer-level nodes, and are passed through another GNN to produce the higher-level node embeddings. In comparison, HGN \citep{fang2020hierarchical} treats the entire hierarchical graph as a whole, by concatenating the precomputed encodings of all levels into the node features, and passing them into a single GNN. There is also a group of simple graph methods \citep{Ying2018DiffPool,Lee2019SAGPool,Gao2019GraphUNets} that propose to learn hierarchical relations in their GNN layers through graph pooling, but they do not use hierarchical graph modelling in their input graph structures. 

\paragraph{Existing Higher-Order Graph Datasets} All above-mentioned hypergraph learning methods are evaluated on hypergraphs constructed from citation networks, such as Cora~\citep{McCallum2000Cora}, CiteSeer~\citep{Getoor2005CiteSeer}, PubMed~\citep{Sen2008PubMed} and DBLP\footnote{\url{https://dblp.org/xml/release/}}. Hyperedges are constructed using either of the two ways: co-citation (i.e., articles are grouped in the same hyperedge if they cite the same article) and co-authorship (i.e., articles are grouped in the same hyperedge if they share the same author). However, hypergraphs constructed from citation networks suffer from two severe disadvantages, making them inadequate evaluation metrics: (1) citation networks are too small in size and are prone to overfitting; and (2) the way hyperedges are constructed leads to substantial overlaps between hyperedges, thereby limiting their effectiveness in capturing multi-node relations. \citet{Chien2022YouAreAllSet} propose several more hypergraph datasets including adaptations of Yelp\footnote{\url{https://www.yelp.com/dataset}}, Walmart~\citep{Amburg2020Walmart} and House~\citep{Chodrow2021House}, but these datasets are still relatively small-scale and have not been widely adopted by the hypergraph learning community.
\section{Hybrid Graph Datasets}

\subsection{Definition of Hybrid Graphs}

Formally, a simple graph $\mathcal{G}=(\mathcal{V},\mathcal{E})$ is a collection of nodes $\mathcal{V}$ and edges $\mathcal{E}\subseteq\mathcal{V}\times\mathcal{V}$ between pairs of nodes. And this graph abstraction assumes that each edge only connects two nodes. However, as discussed in the previous sections, many real-world networks have more complex node relations than just pairwise relations. Attempts made to capture such complex relations include hypergraphs \citep{McCallum2000Cora,Getoor2005CiteSeer,Sen2008PubMed,Chien2022YouAreAllSet}, where a hyperedge can connect more than two nodes, and hierarchical graphs \citep{Lemons2011Hierarchical}, where nodes are organised into multiple levels. A hypergraph $\mathcal{G} = (\mathcal{V}, \mathcal{E})$ is defined by a set of nodes $\mathcal{V}$ and a set of hyperedges $\mathcal{E}$, where edges $e \in \mathcal{E}$ are arbitrary subsets of $\mathcal{V}$. A hierarchical graph is a graph $\mathcal{G} = (\mathcal{V}, \mathcal{E})$ with an acyclic parent function $f_p: \mathcal{V} \to \mathcal{V}$, which defines the node hierarchy. Both hypergraph and hierarchical graph capture more complex node relations than a simple graph, but they still cannot fully capture the relations in real-world applications. Therefore, there is a need for a more general complex graph abstraction that combines simple graphs, hypergraphs and hierarchical graphs.

\begin{figure}[t]
    \centering
    \includegraphics[width=0.8\textwidth]{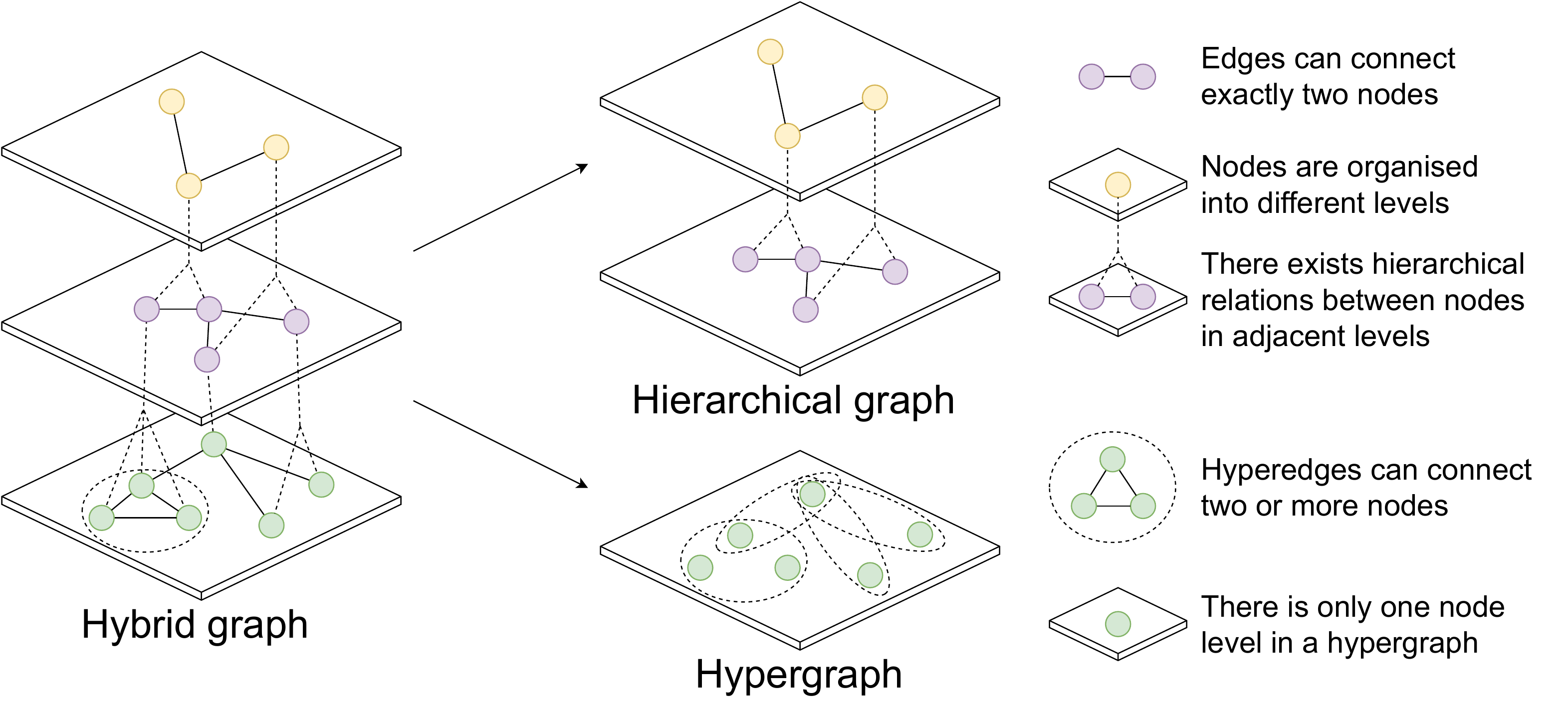}
    \caption{A hybrid graph (left) can (1) contain multiple node levels, where nodes in adjacent levels can have hierarchical relations, and every node at a lower hierarchy must belong to a node at a higher hierarchy; and (2) contain both simple edges (edges that connect exactly two nodes) and hyperedges (edges can connect two or more nodes). It can be transformed into a hierarchical graph (upper right), a hypergraph (lower right) or even a simple graph, by tightening relevant constraints.}
    \label{fig:hybrid-graph}
\end{figure}

We define a \emph{hybrid graph} as $\mathcal{G} = (\mathcal{V}, \mathcal{E}, f_p)$, where $\mathcal{V}$ is the set of nodes, $\mathcal{E}$ is the set of hyperedges, and $f_p: \mathcal{V} \to \mathcal{V}$ is an acyclic parent function. Each hyperedge $e \in \mathcal{E}$ is a non-empty subset of $\mathcal{V}$, which may contain two or more nodes, and is assigned a positive weight $w(e)$. The parent node $f_p(v)$ of a node $v \in \mathcal{V}$ is at the next level up in the node hierarchy. This unified hybrid graph definition encloses the traditional definitions of simple graphs, hypergraphs and hierarchical graphs, making each of them a special type of hybrid graph with some extra constraints:

\begin{itemize}
    \item Simple graph: a hybrid graph is a simple graph if and only if
        \begin{enumerate}
             \item $\forall e \in \mathcal{E}.\:|e|=2$, i.e., a simple graph can only contain simple edges;
             \item $f_p$ is a null function, i.e., a simple graph can only contain a single node level. Therefore, $f_p$ can also be omitted in the simple graph notation.
        \end{enumerate}

    \item Hypergraph: a hybrid graph is a hypergraph if and only if 
        \begin{enumerate}
             \item $\exists e \in \mathcal{E}.\:|e| \geq 3$, i.e., a hypergraph contains at least one hyperedge;
             \item $f_p$ is a null function, i.e., a hypergraph can only contain a single node level. Therefore, $f_p$ can also be omitted in the hypergraph notation.
        \end{enumerate}

    \item Hierarchical graph: a hybrid graph is a hierarchical graph if and only if
        \begin{enumerate}
            \item $\forall e \in \mathcal{E}.\:|e|=2$, i.e., a hierarchical graph can only contain simple edges;
            \item $\exists v \in \mathcal{V}.\: f_p(v)\neq v$, i.e., a hierarchical graph contains at least two node levels;
            \item In a hierarchical graph, except for the nodes at the highest level, every node must be connected to another node in the next level up in the node hierarchy.
        \end{enumerate}
\end{itemize}

Figure~\ref{fig:hybrid-graph} illustrates the properties of a hybrid graph. In graph representation learning, a hybrid graph can be represented as a tuple of features $(\mathbf{X}, \mathbf{E}, \mathbf{H}, \mathbf{W}, \mathbf{R})$.
$\mathbf{X} \in \mathbb{R}^{|\mathcal{V}| \times d_v}$ is the node feature matrix of the hybrid graph, with each row $\mathbf{x}_{v} \in \mathbb{R}^{d_v}$ being the $d_v$-dimensional features of node $v$. 
$\mathbf{E} \in \mathbb{R}^{|\mathcal{E}| \times d_e}$ is the hyperedge feature matrix of the hybrid graph, with each row $\mathbf{e}_{e} \in \mathbb{R}^{d_e}$ being the $d_e$-dimensional features of hyperedge $e$. 
$\mathbf{H} \in \{0,1\}^{|\mathcal{V}| \times |\mathcal{E}|}$ is the node-hyperedge incidence matrix of the hybrid graph, with each entry $H_{ve}=1$ if $v\in e$ and 0 otherwise. We can optionally separate out the simple edges from $\mathbf{H}$ and form an adjacency matrix $\mathbf{A} \in \{0,1\}^{|\mathcal{V}| \times |\mathcal{V}|}$, where $A_{uv}=1$ if $\{u,v\} \in \mathcal{E}$. Note that simple and hierarchical graphs only contain adjacency matrices. 
$\mathbf{W} \in \mathbb{R}^{|\mathcal{E}| \times |\mathcal{E}|}$ is a diagonal matrix containing the weights of the hyperedges, with each on-diagonal entry $W_{ee}=w(e)$. 
$\mathbf{R} \in \{0,1\}^{|\mathcal{V}| \times |\mathcal{V}|}$ is the parent relation matrix of the hybrid graph, with each entry $R_{uv}=1$ if $f_p(u)=v$ and 0 otherwise. If the nodes are arranged in ascending order of their levels in the node hierarchy, then $\mathbf{R}$ becomes an upper-triangular matrix.
Note that $\mathbf{W} = \mathbf{I}_{|\mathcal{E}|}$ and $\mathbf{R} = \mathbf{I}_{|\mathcal{V}|}$ for simple graphs and hypergraphs, therefore they can be omitted in their representations.

\subsection{Dataset Construction}

\begin{figure}[t]
    \centering
    \includegraphics[width=0.8\textwidth]{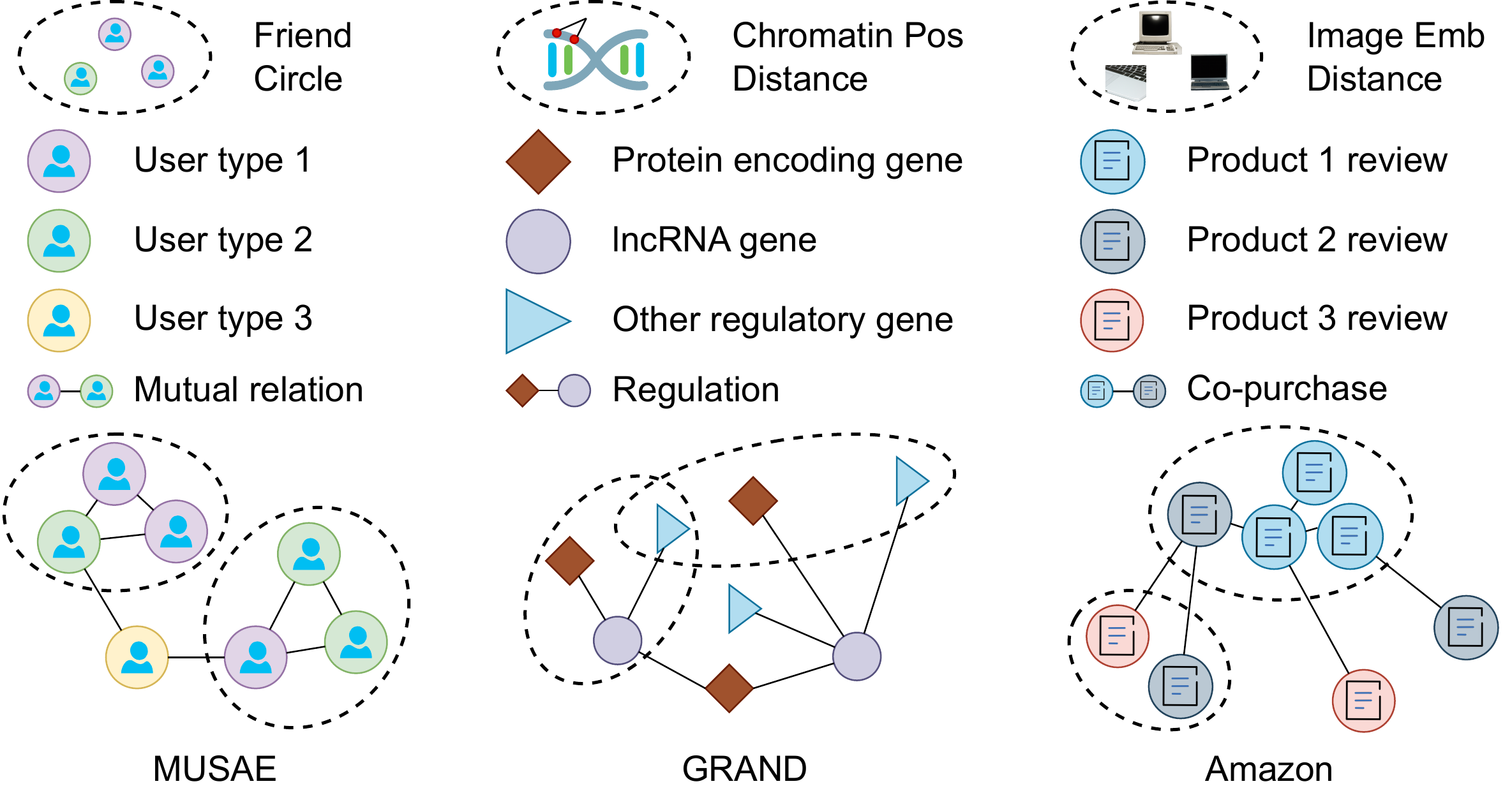}
    \caption{Construction of the HGB datasets. The MUSAE datasets are social and knowledge networks, where the hyperedges are constructed from friend circles or mutually linked page groups. The GRAND datasets are gene regulatory networks, where the hyperedges are formed using the positions of genomic elements on the chromosome. The Amazon datasets are product co-review networks, where the hyperedges are built from the clusters of the image embeddings of the products.}
    \label{fig:datasets}
\end{figure}

The hybrid graph, as a versatile data structure, presents a more comprehensive view of graph data. It not only captures simple pairwise relationships as a simple graph does but also retains complex multi-node interactions similar to hypergraphs and hierarchical graphs. Therefore, it outperforms traditional graph structures in capturing nuanced data relationships. In light of this, we introduce the \emph{Hybrid Graph Benchmark (HGB)}, a novel collection of hybrid graph datasets consisting of 23 hybrid graphs derived from real-world networks across varied domains, including biology, social media, and e-commerce. Care has been taken to ensure that the datasets do not contain any personally identifiable information. Most of these datasets are one-level hybrid graphs containing both simple edges and hyperedges. Meanwhile, if we treat the hyperedges as higher-level virtual nodes representing collections of nodes, then our constructed datasets can also be interpreted as hybrid graphs with a two-level node hierarchy. Figure \ref{fig:datasets} provides an overview of these graphs and their construction mechanisms. Table \ref{table:dataset_stats} reports the key hybrid graph statistics for each dataset group. The HGB datasets can be splitted into three groups according to their construction processes: 

\paragraph{MUSAE} We build eight social networks derived from the Facebook pages, GitHub developers and Twitch gamers, plus three English Wikipedia page-page networks on specific topics (chameleons, crocodiles and squirrels) based on the MUSAE~\citep{Rozemberczki2021MUSAE} datasets. Nodes represent users or articles, and edges are mutual followers relationships between the users, or mutual links between the articles. In addition to the original MUSAE datasets, we construct the hyperedges to be mutually connected sub-groups that contain at least three nodes (i.e., maximal cliques with sizes of at least 3). We also enable each dataset to have an option to use either the raw node features, or the preprocessed node embeddings as introduced in MUSAE.

\paragraph{GRAND} We select and build ten gene regulatory networks in different tissues and diseases from GRAND~\citep{ben2022grand}, a public database for gene regulation. Nodes represent gene regulatory elements~\citep{maston2006transcriptional} with three distinct types: protein-encoding gene, lncRNA gene~\citep{long2017lncrnas}, and other regulatory elements. Edges are regulatory effects between genes. The task is a multi-class classification of gene regulatory elements. We train a CNN~\citep{eraslan2019deep} and use it to take the gene sequence as input and create a 4,651-dimensional embedding for each node. The hyperedges are constructed by grouping nearby genomic elements on the chromosomes, i.e., the genomic elements within 200k base pair distance are grouped as hyperedges. 

\pagebreak 
\paragraph{Amazon} Following existing works on graph representation learning on e-commerce networks~\citep{shchur2018pitfalls,zeng2019graphsaint}, we further build two e-commerce hypergraph datasets based on the Amazon Product Reviews dataset~\citep{mcauley2015image,he2016ups,ni2019justifying}. Nodes represent products, and an edge between two products is established if a user buys these two products or writes reviews for both. However, unlike those existing datasets, we introduce the image modality into the construction of hyperedge. To be specific, the raw images are fed into a CLIP~\citep{radford2021learning} classifier, and a 512-dimensional feature embedding for each image is returned to assist the clustering. The hyperedges are then constructed by grouping products whose image embeddings' pairwise distances are within a certain threshold.

\begin{table}
  \centering
  \caption{Aggregated dataset statistics of HGB.}
  \label{table:dataset_stats}
  \resizebox{\textwidth}{!}{\begin{tabular}{lrrrrrrr}
    \toprule
    Name     & \#Graphs & \makecell[r]{Avg.\\\#Nodes} & \makecell[r]{Avg.\\\#Edges} & \makecell[r]{Avg.\\\#Hyperedges} &\makecell[r]{Avg.\\Node\\Degree} & \makecell[r]{Avg.\\Hyperedge\\Degree} & \makecell[r]{Avg.\\Clustering\\Coef.} \\
    \midrule
     MUSAE-GitHub    & 1 & 37,700 & 578,006 & 223,672 & 30.7 &  4.6 & 0.168 \\
     MUSAE-Facebook  & 1 & 22,470 & 342,004 & 236,663 & 30.4 &  9.9 & 0.360 \\
     MUSAE-Twitch    & 6 &  5,686 & 143,038 & 110,142 & 50.6 &  6.0 & 0.210 \\
     MUSAE-Wiki      & 3 &  6,370 & 266,998 & 118,920 & 88.8 & 14.4 & 0.413 \\
    \midrule
     GRAND-Tissues   & 6 &  5,931 &   5,926 &  11,472 &  2.0 &  1.3 & 0.000 \\
     GRAND-Diseases  & 4 &  4,596 &   6,252 &   7,743 &  2.7 &  1.3 & 0.000 \\
    \midrule
    Amazon-Computers & 1 & 10,226 &  55,324 &  10,226 & 10.8 &  4.0 & 0.249 \\
    Amazon-Photos    & 1 &  6,777 &  45,306 &   6,777 & 13.4 &  4.8 & 0.290 \\
    \bottomrule
  \end{tabular}}
\end{table}
\section{Evaluation Framework}

\subsection{Overview}

\begin{figure}[t]
    \centering
    \includegraphics[width=\textwidth]{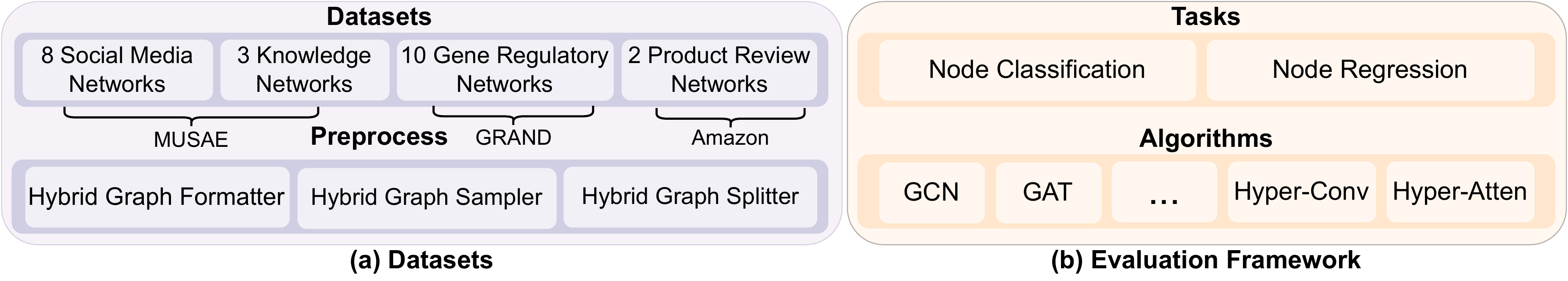}
    \caption{An overview of the HGB framework. (a) The HGB datasets include 23 real-world hybrid graphs, which preserve both simple and higher-order graph structures. (b) We also build a lightweight training-test framework using PyTorch Lightning and PyTorch Geometric.}
    \label{fig:framework}
\end{figure}

We create an extensible evaluation framework in HGB, simplifying the process of training and assessing GNNs for both simple graphs and hypergraphs. Figure \ref{fig:framework} illustrates the key components of HGB. 23 hybrid graphs are used to train and evaluate the seven GNNs, including four GNNs for simple graphs: GCN~\citep{kipf2016semi}, GraphSAGE~\citep{hamilton2017inductive}, GAT~\citep{velivckovic2017graph}, GATv2~\citep{brody2021attentive}; two hypergraph GNNs: HyperConv, HyperAtten~\citep{Bai2021HyperConvAtten}; and one sampling-based training strategy: GraphSAINT~\citep{zeng2019graphsaint}. To fairly compare these models, we evaluate them under the same hyperparameter settings, which are listed in Appendix~\ref{appendix:details}. We repeat each experiment 5 times with different random seeds, and report their means and standard deviations in Appendix~\ref{appendix:results}. The experiments are evaluated with the accuracy for the node classification task and the mean square error (MSE) for node regression tasks. Our results highlight the challenges and research gaps in developing effective GNNs for real-world hybrid graphs:

\begin{itemize}
    \item Existing hypergraph GNNs may not outperform simple graph GNNs, even when the hyperedges provide meaningful information to the task. (Section~\ref{section:simple-vs-hyper-gnns})
    \item The performance of conventional hypergraph GNNs can be improved with graph samplers that samples the node mainly based on the simple edge information. (Section~\ref{section:sampler-study})
    \item Combining both simple and higher-order graph information can be substantially beneficial for node classification and regression tasks in higher-order graphs. (Section~\ref{section:intergrate-simple-hyper-info})
\end{itemize}

\subsection{Assessing Performance: Hypergraph GNNs vs. Simple Graph GNNs on HGB} \label{section:simple-vs-hyper-gnns}

We report the mean accuracies of four simple graph GNNs (GCN, GraphSAGE, GAT, GATv2) and two hypergraph GNNs (HyperConv, HyperAtten) on HGB in Appendix \ref{appendix:results}. For MUSAE and GRAND, the information embedded in the hypergraph space and simple graph space do not contribute to the task objective. For the Amazon datasets, since the hyperedges are constructed using the embedding of actual product images, nodes within a hyperedge can provide meaningful information to the task. Therefore, we expect hypergraph GNNs to outperform simple graph GNNs on Amazon. However, our experiments show that the performance gain from hypergraph GNNs on Amazon is only marginal.

\begin{figure}[t]
    \centering
    \begin{subfigure}[t]{0.503\textwidth}
        \includegraphics[width=\columnwidth]{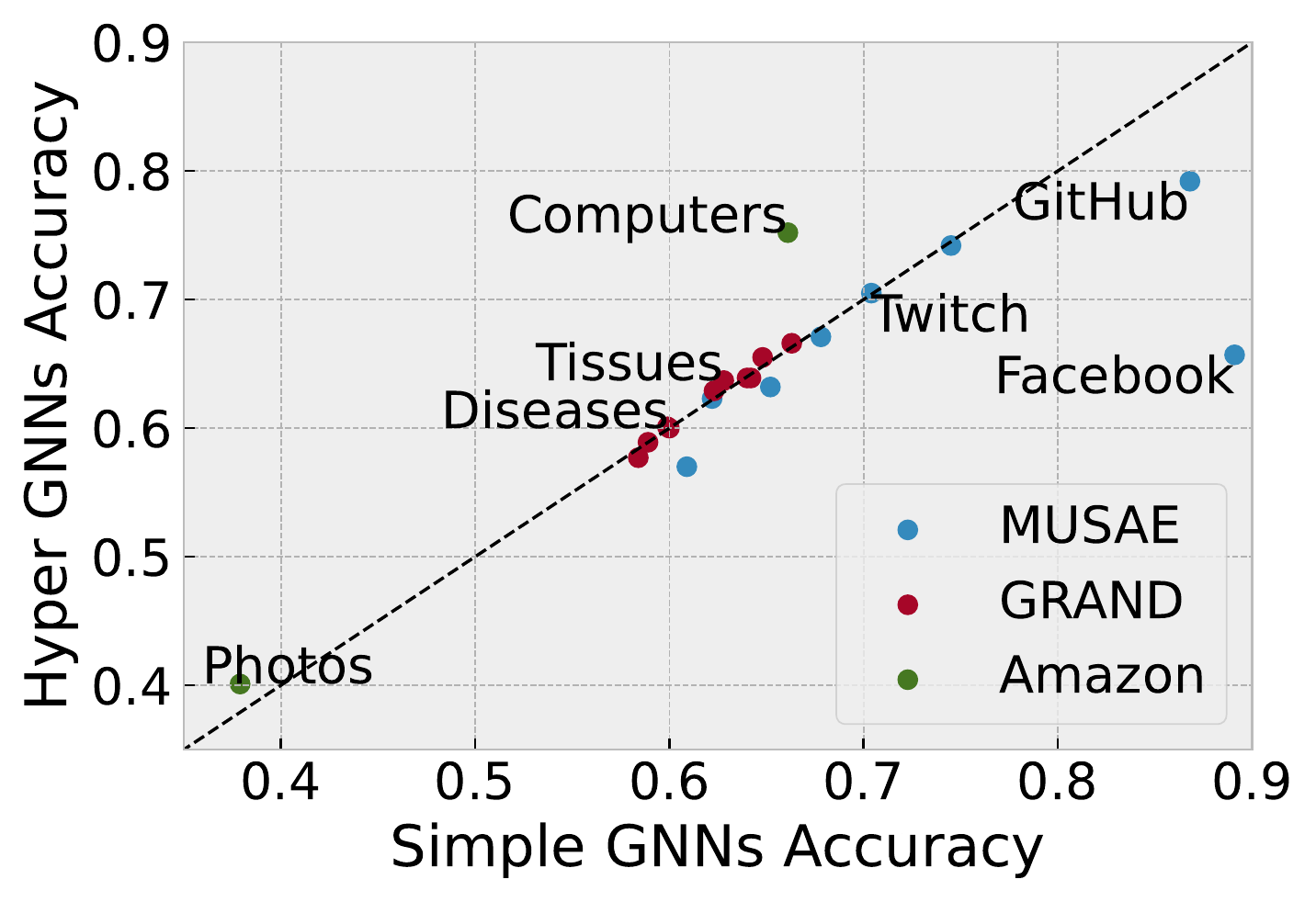}
        \vskip -0.5\baselineskip
        \caption{}
        \vskip -0.5\baselineskip
        \label{fig:eval_compare_scatter}
    \end{subfigure}
    \begin{subfigure}[t]{0.488\textwidth}
        \includegraphics[width=\columnwidth]{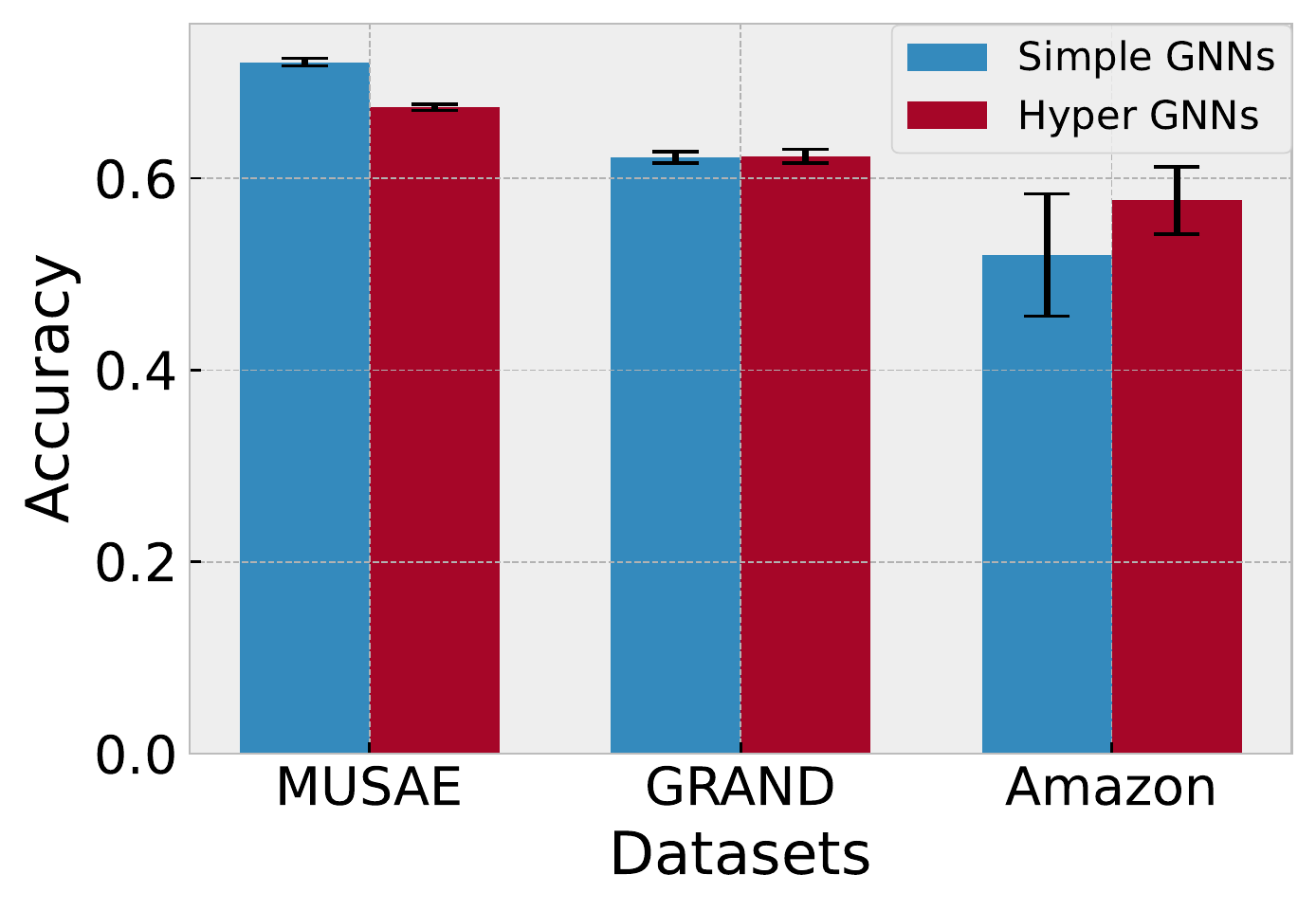}
        \vskip -0.5\baselineskip
        \caption{}
        \vskip -0.5\baselineskip
        \label{fig:eval_compare_bar}
    \end{subfigure}
    \caption{Evaluating the accuracy of simple-graph and hypergraph GNNs on HGB: (a) The scatter plot of the accuracies of hypergraph GNNs with respect to simple graph GNNs on each dataset. The black dashed line at $y=x$ serves as a reference line where both GNN types perform equally well, while a dot above this reference line means hypergraph GNNs perform better than simple graph GNNs on a given node classification task, and vice versa. The plot indicates that hypergraph GNNs match simple graph GNNs' accuracies on GRAND, underperform on MUSAE, and outperform on Amazon. (b) The bar chart of the accuracies of both GNN types aggregated within MUSAE, GRAND, and Amazon, shows that the performance improvement of hypergraph GNNs on Amazon compared to simple graph GNNs may not be significant, after taking the standard deviations into consideration.}
    \label{fig:eval_compare}
\end{figure}

Figure \ref{fig:eval_compare_scatter} shows a pairwise comparison of two types of GNNs. In both MUSAE and GRAND, hypergraph GNNs perform equally or less well than the simple graph methods, while they perform better than simple graph GNNs on two Amazon Review graphs. However, Figure \ref{fig:eval_compare_bar}, the aggregated mean accuracy across all graphs with the standard deviation, indicates that the performance gain of using hypergraph GNNs may not be significant. 

\subsection{Optimised Sampling in Hybrid Graph}
\label{section:sampler-study}

Various hypergraph sampling strategies are proposed for sampling subgraphs with the purpose of preserving the graph statistics~\citep{choe2022midas,dyer2021sampling}. However, there is a lack of practical implementations of hypergraph samplers and evaluations of their efficacy. Following the work by GraphSAINT \citep{zeng2019graphsaint} on simple graph sampling, we propose \emph{HybridGraphSAINT}, a class of hybrid graph samplers employing GraphSAINT's graph sampling approaches. In HybridGraphSAINT, we adopt the same sampling strategies in GraphSAINT for sampling the simple graph components in a hybrid graph, making three different types of samplers: node sampler (HybridGraphSAINT-Node), edge sampler (HybirdGraphSAINT-Edge), and random walk sampler (HybridGraphSAINT-RW). As for the hyperedges, we use an intuitive procedure that any hyperedges containing at least one node in the sampled subgraph are retained, but all nodes not in the subgraph are masked out from those hyperedges. By using this method, we preserve the original hybrid graph characteristics in the sampled subgraphs to the maximum extent. We also construct two na\"{i}ve random samplers as baselines for evaluation: random node sampler and random hyperedge sampler, which randomly sample a subset of nodes/hyperedges from the original hybrid graph according to a uniform sampling distribution. However, subgraphs sampled using the random node sampler can be very sparse, while subgraphs sampled using the random hyperedge sampler can be very dense. 

We evaluate these samplers on MUSAE-GitHub and MUSAE-Facebook, the two largest graphs in HGB. Firstly, we sample subgraphs with various samplers to examine how they preserve the structure of the original graphs. This is measured by three graph statistics: average node/hyperedge degree, and clustering coefficient. We sample subgraphs multiple times and report the average graph statistics in Table \ref{table:subgraphStats} of Appendix \ref{appendix:results}. Among all constructed samplers, HybirdGraphSAINT-Edge and HybirdGraphSAINT-RW perform the best in preserving the graph-level statistics. 

\begin{figure}[t]
    \centering
    \begin{subfigure}[t]{0.505\textwidth}
        \includegraphics[width=\columnwidth]{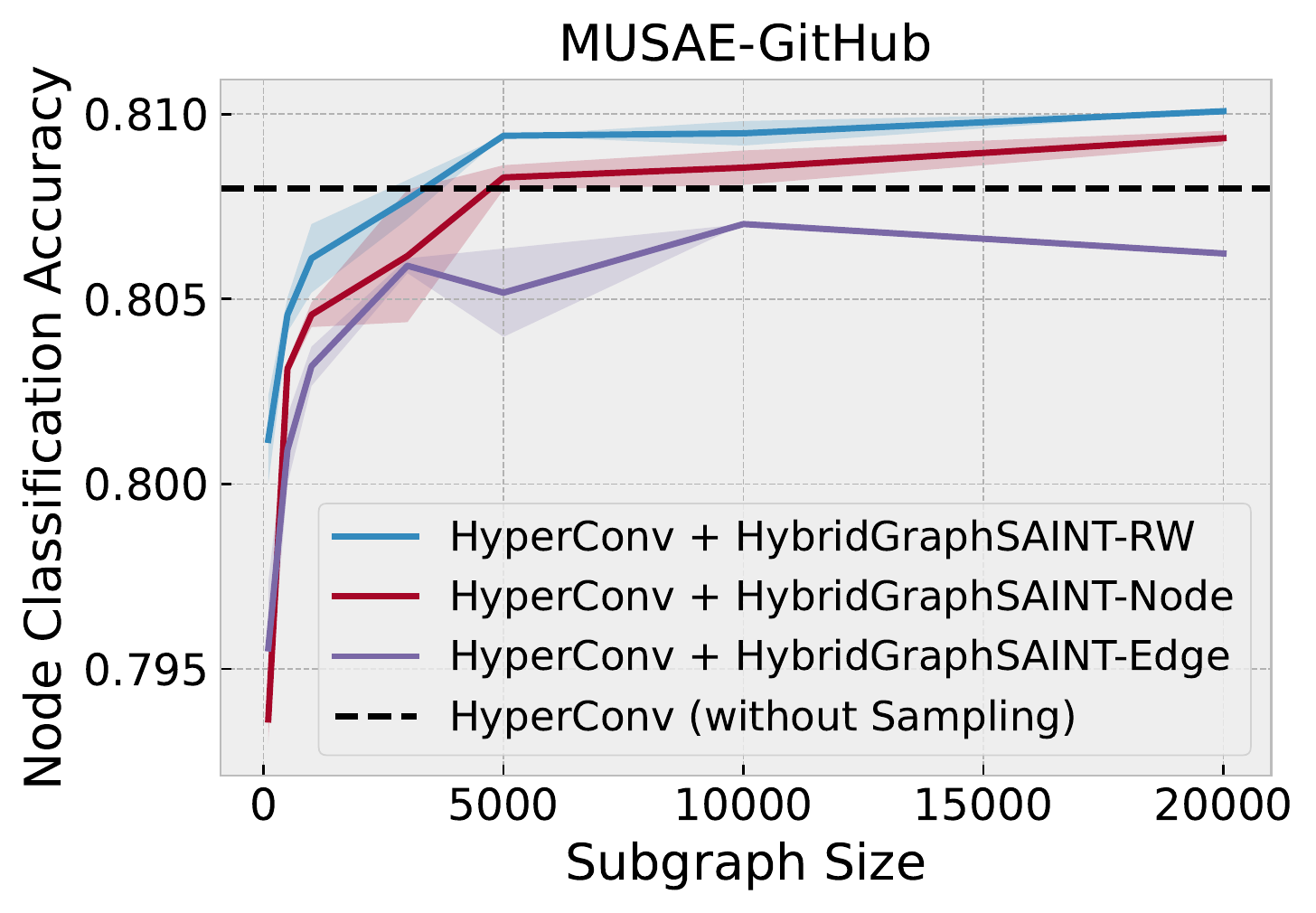}
        \vskip -0.5\baselineskip
        \caption{}
    \end{subfigure}
    \begin{subfigure}[t]{0.484\textwidth}
        \includegraphics[width=\columnwidth]{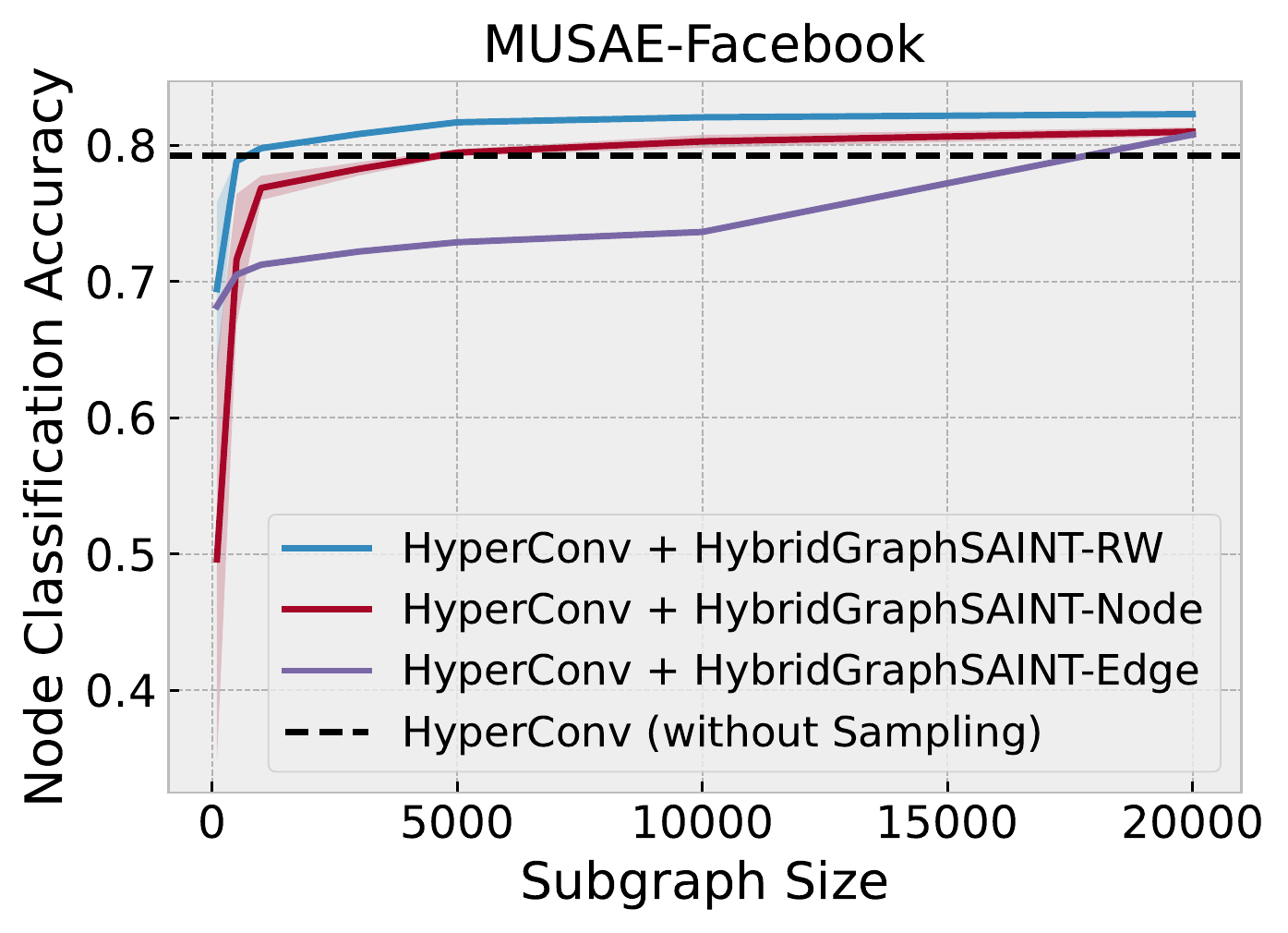}
        \vskip -0.5\baselineskip
        \caption{}
    \end{subfigure}
    \vspace*{-0.5\baselineskip}
    \caption{Node classification accuracy of different sampling techniques on (a) MUSAE-GitHub and (b) MUSAE-Facebook. The charts compare the accuracy of HyperConv model with three variants of HybirdGraphSAINT samplers at different subgraph sizes. The size of the subgraph is measured by the number of nodes.  The black dashed line indicates the performance of HyperConv trained on the whole graph without any sampling. The accuracy increases as subgraph sizes increased, with both HybridGraphSAINT-RW and HybridGraphSAINT-Node outperforming standard HyperConv. HybirdGraphSAINT-RW consistently shows superior performance across different sampling sizes.}
    \vspace*{-0.5\baselineskip} 
    \label{fig:eval_sampler}
\end{figure}

We also evaluate HyperConv performance when paired with three different HybridGraphSAINT samplers. Figure \ref{fig:eval_sampler} shows distinct patterns regarding how the accuracy varied with different subgraph sizes. Evaluations were carried out on two distinct datasets, yielding the following key observations (1) There exists a positive correlation between the size of sampled subgraphs and model accuracy across both datasets. (2) At subgraph sizes of 5000 or larger, both HybridGraphSAINT-RW and HybridGraphSAINT-Node outperform vanilla HyperConv (trained on the whole graph without sampling). This highlights the potential for improved performance of hypergraph GNNs when integrating our HybridGraphSAINT samplers. This performance gain could be explained by the reduction of over-smoothing brought by sampling subgraphs.

Overall, the HybridGraphSAINT samplers, which mostly use simple edge information, can both effectively preserve hypergraph statistics and improve the performance of the hypergraph learning algorithms. This again underpins the advantage of hybrid graphs over conventional hypergraphs and simple graphs, emphasising the need to preserve multi-level information.

\subsection{Integrating Simple Graph and Hypergraph Information}
\label{section:intergrate-simple-hyper-info}


To test whether combining simple and higher-order graph information could improve the performance of GNNs, we propose a simple algorithm called \emph{Linear Probe Graph Neural Networks (LP-GNNs)}. LP-GNN consists of two GNNs $f_{\text{GNN}_1}$, $f_{\text{GNN}_2}$, plus a linear layer $f(\mathbf{x}) = \bm{\theta}\mathbf{x}+\mathbf{b}$. For a given hybrid graph $\mathcal{G} = (\mathbf{X}, \mathbf{E}, \mathbf{H}, \mathbf{W}, \mathbf{R})$, where $\mathbf{x}$ denotes the input features of a node, LP-GNN is defined as
\begin{equation}
    \text{LP-GNN}(\mathbf{x}) = \bm{\theta}\cdot\mathrm{CONCATENATE}(f_{\text{GNN}_1}(\mathbf{x}),f_{\text{GNN}_2}(\mathbf{x})) + \mathbf{b}
    \label{eq:eq1}
\end{equation}
For node regression tasks, $\text{LP-GNN}(\mathbf{x})$ is directly used as the final output. For node classification tasks, a log softmax function is applied to $\text{LP-GNN}(\mathbf{x})$ to produce the final output. 

\Cref{Table:lp-gnn} shows the performance of three LP-GNNs with the GNN pair being GCN+GAT, GCN+HyperConv, and GAT+HyperConv. These models are benchmarked across five different HGB datasets, and their performances are contrasted with those of simple graph GNNs. Our focus lies on the performance of LP-GCN+HyperConv and LP-GAT+HyperConv, which combine simple edge and hyperedge information. In three of the datasets (Wiki-Chameleon, Breast, and Leukemia), LP-GAT+HyperConv and LP-GCN+HyperConv demonstrate the highest efficacy. In GitHub and Computers, those two models also tend to match other GNNs' performances. LP-GNNs employ a very simple strategy to combine information from two levels and show its advantage. This evidents the potential advantages of integrating information across multiple levels.

\begin{table}
    \small
    \caption[Accuracies of LP-GNNs, GCN, GAT, and HyperConv]{Accuracies of LP-GNNs, GCN, GAT, and HyperConv. The reciprocals of mean square errors (1/MSE) are reported for the GNNs' node regression task performances on the Chameleon dataset.}
    \label{Table:lp-gnn}
    \centering
    \resizebox{\textwidth}{!}{\begin{tabular}{llllll}
        \toprule
        Method & GitHub & Chameleon & Breast & Leukemia & Computers \\
        \midrule
        RandomGuess & 0.250 & --- & 0.333 & 0.333 & 0.100 \\
        \midrule
        GCN         & \textbf{0.872 $\pm$ 0.000} & 0.137 $\pm$ 0.000 & 0.639 $\pm$ 0.010 & 0.582 $\pm$ 0.001 & 0.756 $\pm$ 0.041 \\ 
        GAT         & 0.864 $\pm$ 0.001 & 0.152 $\pm$ 0.004 & 0.643 $\pm$ 0.001 & 0.587 $\pm$ 0.005 & 0.742 $\pm$ 0.043 \\
        HyperConv   & 0.808 $\pm$ 0.001 & 0.138 $\pm$ 0.000 & 0.645 $\pm$ 0.001 & 0.586 $\pm$ 0.003 & 0.842 $\pm$ 0.020 \\
        \midrule
        LP-GCN+GAT  & 0.867 $\pm$ 0.001 & 0.181 $\pm$ 0.009 & 0.626 $\pm$ 0.001 & 0.590 $\pm$ 0.002 & \textbf{0.930 $\pm$ 0.000} \\
        LP-GCN+HyperConv & \textbf{0.872 $\pm$ 0.000} & 0.181 $\pm$ 0.001 & 0.652 $\pm$ 0.006 & \textbf{0.604 $\pm$ 0.004} & 0.913 $\pm$ 0.001 \\ 
        LP-GAT+HyperConv & 0.860 $\pm$ 0.002 & \textbf{0.205 $\pm$ 0.002} & \textbf{0.657 $\pm$ 0.001} & 0.601 $\pm$ 0.002 & 0.930 $\pm$ 0.007 \\
        \bottomrule
    \end{tabular}}
    \vspace*{-0.5\baselineskip}
\end{table}
\section{Conclusion and Discussion}

We introduce the concept of \emph{hybrid graphs}, a unified representation of various complex graph structures. Additionally, we present the \emph{Hybrid Graph Benchmark (HGB)}, a collection of real-world hybrid graph datasets accompanied with an extensible GNN evaluation framework. Through extensive experiments, we demonstrate that existing hypergraph GNNs are not guaranteed to outperform simple graph GNNs on large-scale complex networks. In light of this, we propose a simple model called \emph{Linear Probe Graph Neural Networks (LP-GNNs)}, which integrates simple graph and hypergraph information, and leverages the node prediction performance on the hybrid graphs. We believe that HGB can significantly aid the research in complex graph representation learning, and provide valuable insights to future research directions for the graph representation learning community.

\paragraph{Limitations} We define hybrid graphs as graphs with multiple levels. However, the current collection of datasets in HGB primarily consists of shallow hybrid graphs with a two-level node hierarchy, and networks with deeper node hierarchies are desired. In MUSAE, hyperedges are constructed using the maximal cliques of a graph, which might lead to an overlap of information between the simple edges and hyperedges. In GRAND, the majority of gene regulatory graphs exhibit bipartite-like characteristics, which may be limited in connectivity and clustering for hyperedge constructions. In Amazon, hyperedges are created based on a pre-defined threshold. Our preliminary experiments show that the selection of this threshold value can influence the structure of the resultant hybrid graphs, and a more thorough investigation is required to find the range of the most appropriate threshold values.

\paragraph{Future Work} HGB will be updated on a regular basis, and we welcome inputs from the community. In the future versions of HGB, we plan to to enable multiple versions of Amazon datasets by adjusting various pairwise distance thresholds, and include more real-world large-scale networks featuring multi-level node hierarchies. Additionally, as simply concatenating simple and hypergraph information followed by a linear transformation can already enhance the node prediction performance on hybrid graphs, we will also seek to further improve the performance following this idea, by incorporating more fine-grained operations to combine simple graph and higher-order graph information.


\begin{ack}
This work was performed using the Sulis Tier 2 HPC platform hosted by the Scientific Computing Research Technology Platform at the University of Warwick, and the JADE Tier 2 HPC facility. Sulis is funded by EPSRC Grant EP/T022108/1 and the HPC Midlands+ consortium. JADE is funded by EPSRC Grant EP/T022205/1. Zehui Li acknowledges the funding from the UKRI 21EBTA: EB-AI Consortium for Bioengineered Cells \& Systems (AI-4-EB) award, Grant BB/W013770/1. Xiangyu Zhao acknowledges the funding from the Imperial College London Electrical and Electronic Engineering PhD Scholarship. Mingzhu Shen acknowledges the funding from the Imperial College London President's PhD Scholarship. For the purpose of open access, the authors have applied a Creative Commons Attribution (CC BY) licence to any Author Accepted Manuscript version arising.
\end{ack}

\bibliographystyle{apalike}
\bibliography{references}


\newpage
\appendix
\section{Full Statistics of HGB Datasets}

\begin{table}[h]
    \centering
    \caption{Statistics of all 23 hybrid graph datasets in HGB.}
    \label{table:all-graph-stats}
    \resizebox{\textwidth}{!}{\begin{tabular}{lrrrrrrr}
        \toprule
        Name    & \makecell[r]{\#Nodes} & \makecell[r]{\#Edges} & \makecell[r]{\#Hyperedges} & \makecell[r]{Avg.\\Node\\Degree} & \makecell[r]{Avg.\\Hyperedge\\Degree} & \makecell[r]{\#Node\\Features} & \makecell[r]{\#Classes} \\
        \midrule
        MUSAE-Github & 37,700 & 578,006 & 223,672 & 30.66 & 4.591 & 4,005 or 128 & 4 \\ 
        MUSAE-Facebook & 22,470 & 342,004 & 236,663 & 30.44 & 9.905 & 4,714 or 128 & 4 \\
        MUSAE-Twitch-DE & 9,498 & 306,276 & 297,315 & 64.49 & 7.661 & 3,170 or 128 & 2 \\ 
        MUSAE-Twitch-EN & 7,126 & 70,648 & 13,248 & 19.83 & 3.666 & 3,170 or 128 & 2 \\  
        MUSAE-Twitch-ES & 4,648 & 118,764 & 77,135 & 51.10 & 5.826 & 3,170 or 128 & 2 \\ 
        MUSAE-Twitch-FR & 6,549 & 225,332 & 172,653 & 68.81 & 5.920 & 3,170 or 128 & 2 \\  
        MUSAE-Twitch-PT & 1,912 & 62,598 & 74,830 & 65.48 & 7.933 & 3,170 or 128 & 2 \\  
        MUSAE-Twitch-RU & 4,385 & 74,608 & 25,673 & 34.03 & 4.813 & 3,170 or 128 & 2 \\  
        MUSAE-Wiki-Chameleon & 2,277 & 62,742 & 14,650 & 55.11 & 7.744 & 3,132 or 128 & Regression \\ 
        MUSAE-Wiki-Crocodile & 11,631 & 341,546 & 121,431 & 58.73 & 4.761 & 13,183 or 128 & Regression \\ 
        MUSAE-Wiki-Squirrel & 5,201 & 396,706 & 220,678 & 152.55 & 30.735 & 3,148 or 128 & Regression \\
        \midrule
        GRAND-ArteryAorta & 5,848 & 5,823 & 11,368 & 1.991& 1.277 &4,651&3\\ 
        GRAND-ArteryCoronary & 5,755 & 5,722 & 11,222 & 1.989& 1.273 &4,651&3\\ 
        GRAND-Breast & 5,921 & 5,910 & 11,400 & 1.996 & 1.281 &4,651&3\\ 
        GRAND-Brain & 6,196 & 6,245 & 11,878 & 2.016 & 1.296 &4,651&3\\ 
        GRAND-Lung & 6,119 & 6,160 & 11,760 & 2.013 & 1.291 &4,651&3\\ 
        GRAND-Stomach & 5,745 & 5,694 & 11,201 & 1.982 & 1.274 &4,651&3\\
        GRAND-Leukemia & 4,651 & 6,362 & 7,812 & 2.736 & 1.324 &4,651&3\\ 
        GRAND-Lungcancer & 4,896 & 6,995 & 8,179 & 2.857 & 1.334 &4,651&3\\
        GRAND-Stomachcancer & 4,518 & 6,051 & 7,611 & 2.679 & 1.312 &4,651&3\\
        GRAND-KidneyCancer & 4,319 & 5,599 & 7,369 & 2.593 & 1.297 &4,651&3\\ 
        \midrule
        Amazon-Computers & 10,226 & 55,324 & 10,226 & 10.82 & 3.000 & 1,000 & 10\\ 
        Amazon-Photos & 6,777 & 45,306 & 6,777 & 13.37 & 4.800 & 1,000 & 10\\ 
        \bottomrule
    \end{tabular}}
\end{table}
\section{Experimental Details} \label{appendix:details}

\subsection{Training Details}

We run all the experiments on NVIDIA A100 PCIe GPU with 40GB RAM (Sulis) and NVIDIA V100 NVLink GPU with 32GB RAM (JADE), with each experiment taking less than 2 minutes. Adam~\citep{kingma2014method} is used as the optimiser, and CosineAnnealingLR~\citep{gotmare2018closer} is used as the learning rate scheduler for all training. All models are trained for 50 epochs. For each experiment, the nodes of the used hybrid graph are split into the train, validation, and test sets with a split ratio of 6:2:2. For node classification tasks, BCEWithLogitsLoss is used as the loss function, which is defined as:

\begin{equation}
    \mathcal{L}_\text{BCEWithLogits}(\mathbf{y},\hat{\mathbf{y}}) = -\frac{1}{n} \sum_{i=1}^{n}\big[y_i \cdot \log(\sigma(\hat{y}_i)) + (1 - y_i) \cdot \log(1 - \sigma(\hat{y}_i))\big]
\end{equation}

where $n$ is the total number of elements in $\mathbf{y}$ and $\hat{\mathbf{y}}$, $y_i$ is the $i$-th element of $\mathbf{y}$, the batch of true values, and $\hat{y}_i$ is the $i$-th element of $\hat{\mathbf{y}}$, the batch of raw (i.e., non-sigmoid-transformed) predicted values. $\sigma$ denotes the sigmoid function, which transforms the raw predictions into the range (0, 1). For node regression tasks, MSELoss is used as the loss function, which is defined as:

\begin{equation}
    \mathcal{L}_\text{MSE}(\mathbf{y},\hat{\mathbf{y}}) = \frac{1}{n} \sum_{i=1}^{n} (y_i - \hat{y}_i)^2
\end{equation}

where $n$ is the total number of elements in $\mathbf{y}$ and $\hat{\mathbf{y}}$, $y_i$ is the $i$-th element of $\mathbf{y}$, the batch of true values, and $\hat{y}_i$ is the $i$-th element of $\hat{\mathbf{y}}$, the batch of predicted values.

\subsection{Hyperparameter Settings}

We perform a hyperparameter search for the learning rate and keep the hidden layer dimension the same for different models, the hyperparameters used for training each architecture are listed in \Cref{table:hyperparams}. All seven GNNs (GCN, GraphSAGE, GAT, GATv2, HyperConv, HyperAtten, and GraphSAINT) share the same learning rate, hidden dimension, and dropout rate. HyperAtten has an additional hyperparameter, which is the hyperedge aggregation function. This function determines how the hyperedge is constructed from the nodes within it. The possible options for this function are `sum' and `concatenate'.  In this work, we have selected `sum' as the hyperedge aggregation function. For GraphSAINT, the additional hyperparameters are subgraph size, measured by the number of nodes in the subgraph, and the batch size, which is the number of subgraphs to sample in each epoch. Different subgraph sizes are applied according to the sizes of the original hybrid graphs.

\begin{table}[h]
    \caption{Hyperparameter selections for the experiments.}
    \label{table:hyperparams}
    \centering
    \begin{tabular}{llc}
        \toprule
        Method & Hyperparameter & Value \\
        \midrule
        \multirow{3}{*}{All} & Learning rate & 0.01 \\
        & Hidden dimension & 32 \\
        & Dropout rate & 0.5 \\
        \midrule
        HyperAtten & Hyperedge aggregation function & Sum \\
        \midrule
        \multirow{4}{*}{GraphSAINT} & Subgraph size (MUSAE-GitHub, MUSAE-Facebook) & 5000 \\
        & Subgraph size (MUSAE-Twitch-PT) & 1000 \\
        & Subgraph size (others) & 3000 \\
        & Batch size & 5 \\
        \bottomrule
    \end{tabular}
\end{table}

\section{Results} \label{appendix:results}
We evaluate the performance of seven GNNs on all 23 HGB datasets. Each experiment is repeated five times with different random seeds, and the results are summarised in \Cref{table:facebook-github,table:twitch,table:grand-tissue,table:grand-disease,table:amazon,table:wiki}. In the node classification tasks of MUSAE, GCN and GraphSAGE perform the best in the on GitHub and Facebook, as shown in \Cref{table:facebook-github}, while all GNNs perform roughly the same on Twitch, as shown in \Cref{table:twitch}. In the three node regression tasks on MUSAE-Wiki, GraphSAINT stands out among other GNNs, as shown in \Cref{table:wiki}. \Cref{table:grand-tissue} shows that HyperConv and HyperAtten outperform other simple graph GNNs on five of the six hybrid graphs in GRAND-Tissues. For hybrid graphs in GRAND-Diseases as shown in \Cref{table:grand-disease}, GraphSAGE exhibits a superior performance. For the two Amazon hybrid graph datasets, as shown in \Cref{table:amazon}, GraphSAINT consistently achieves the best performance. Overall, hypergraph GNNs tend to outperform simple graph GNNs on GRAND-Tissues and Amazon, perform equally as simple graph GNNs on MUSAE-Twitch, MUSAE-Wiki and GRAND-Diseases, and underperform simple graph GNNs on MUSAE-GitHub and MUSAE-Facebook.

\Cref{table:subgraphStats} show the statistics of the subgraphs obtained by five different samplers. We sample subgraphs from both MUSAE-GitHub and MUSAE-Facebook, and then compute the average statistics of the sampled subgraphs. HybridGraphSAINT-Edge and HybridGraphSAINT-RW perform the best in preserving the graph-size agnostic statistics. The two random samplers tend to sample subgraphs with distinct structures from the original hybrid graphs.

\begin{table}[htbp]
    \caption{Accuracies of the selected GNNs on the MUSAE-Facebook and MUSAE-GitHub datasets. GCN and GraphSAGE perform the best on these datasets, and the hypergraph GNNs tend to perform worse than the simple graph GNNs.}
    \label{table:facebook-github}
    \centering
    \begin{tabular}{lll}
        \toprule
        Method     & Facebook     & GitHub\\
        \midrule
        RandomGuess & 0.250   & 	0.250  \\
        \midrule
        GCN        & 0.886 $\pm$ 0.001 & \textbf{0.872 $\pm$ 0.000} \\
        GraphSAGE  & \textbf{0.902 $\pm$ 0.002} & 0.871 $\pm$ 0.002 \\
        GAT        & 0.876 $\pm$ 0.001 & 0.864 $\pm$ 0.001 \\
        GATv2      & 0.901 $\pm$ 0.001 & 0.866 $\pm$ 0.001 \\
        HyperConv  & 0.792 $\pm$ 0.001 & 0.808 $\pm$ 0.001 \\
        HyperAtten & 0.523 $\pm$ 0.002 & 0.775 $\pm$ 0.001  \\
        GraphSAINT & 0.896 $\pm$ 0.001 & 0.871 $\pm$ 0.001  \\
        \bottomrule
    \end{tabular}
\end{table}

\begin{table}[htbp]
    \caption{Accuracies of the selected GNNs on the MUSAE-Twitch datasets. All selected GNNs achieve similar performance on this group of datasets.}
    \label{table:twitch}
    \centering
    \resizebox{\textwidth}{!}{\begin{tabular}{lllllll}
        \toprule
        Method & TwitchES & TwitchFR & TwitchDE & TwitchEN & TwitchPT & TwitchRU\\
        \midrule
        RandomGuess &  0.500 & 0.500 & 0.500 & 0.500 & 0.500 & 0.500 \\
        \midrule
        GCN         & \textbf{0.721 $\pm$ 0.004} & 0.624 $\pm$ 0.001    & 0.655 $\pm$ 0.002  & \textbf{0.620 $\pm$ 0.003} & 0.689 $\pm$ 0.006 & 0.745 $\pm$ 0.000 \\ 
        GraphSAGE   & 0.690 $\pm$ 0.002 & 0.616 $\pm$ 0.003 & \textbf{0.657 $\pm$ 0.001} & 0.605 $\pm$ 0.000 & 0.672 $\pm$ 0.013 & 0.745 $\pm$ 0.001 \\
        GAT         & 0.694 $\pm$ 0.002 & 0.623 $\pm$ 0.000 & 0.645 $\pm$ 0.004 & 0.594 $\pm$ 0.006 & 0.664 $\pm$ 0.007 & 0.743 $\pm$ 0.002 \\
        GATv2       & 0.710 $\pm$ 0.003 & \textbf{0.625 $\pm$ 0.001} & 0.651 $\pm$ 0.003 & 0.618 $\pm$ 0.005 & 0.687 $\pm$ 0.009 & 0.745 $\pm$ 0.000 \\
        HyperConv   & 0.715 $\pm$ 0.001 & 0.624 $\pm$ 0.002 & 0.654 $\pm$ 0.002 & 0.587 $\pm$ 0.007 & \textbf{0.701 $\pm$ 0.005} & 0.741 $\pm$ 0.001 \\
        HyperAtten  & 0.695 $\pm$ 0.000 & 0.623 $\pm$ 0.001 & 0.610 $\pm$ 0.003 & 0.553 $\pm$ 0.003 & 0.641 $\pm$ 0.000 & 0.743 $\pm$ 0.000 \\
        GraphSAINT  & 0.713 $\pm$ 0.008 & 0.622 $\pm$ 0.003 & 0.653 $\pm$ 0.004 & 0.610 $\pm$ 0.011 & 0.677 $\pm$ 0.006 &	\textbf{0.746 $\pm$ 0.002} \\
        \bottomrule
    \end{tabular}}
\end{table}

\begin{table}[htbp]
    \small
    \caption{Accuracies of the selected GNNs on the GRAND-Tissues datasets. Hypergraph GNNs~tend~to outperform simple graph GNNs on these datasets, with HyperAtten being the best-performing method.}
    \label{table:grand-tissue}
    \centering
    \resizebox{\textwidth}{!}{\begin{tabular}{lllllll}
        \toprule
        Method & ArteryAorta & ArteryCoronary & Breast & Brain & Lung & Stomach \\
        \midrule
        RandomGuess & 0.333 & 0.333 & 0.333 & 0.333 & 0.333 & 0.333\\
        \midrule
        GCN         & 0.627 $\pm$ 0.007 & 0.662 $\pm$ 0.001 & 0.639 $\pm$ 0.010 & 0.625 $\pm$ 0.000 & 0.650 $\pm$ 0.000 & \textbf{0.643 $\pm$ 0.000} \\ 
        GraphSAGE   & 0.628 $\pm$ 0.002 & 0.663 $\pm$ 0.001 & 0.644 $\pm$ 0.000 & 0.618 $\pm$ 0.002 & 0.646 $\pm$ 0.005 & 0.630 $\pm$ 0.010 \\
        GAT         & 0.628 $\pm$ 0.004 & 0.663 $\pm$ 0.000 & 0.643 $\pm$ 0.001 & 0.625 $\pm$ 0.001 & 0.648 $\pm$ 0.004 & 0.643 $\pm$ 0.000 \\
        GATv2       & 0.630 $\pm$ 0.000 & 0.663 $\pm$ 0.000 & 0.644 $\pm$ 0.000 & 0.624 $\pm$ 0.001 & 0.650 $\pm$ 0.001 & 0.642 $\pm$ 0.000 \\
        HyperConv   & 0.626 $\pm$ 0.008 & 0.662 $\pm$ 0.000 & \textbf{0.645 $\pm$ 0.001} & 0.625 $\pm$ 0.000 & 0.650 $\pm$ 0.000 & 0.643 $\pm$ 0.000 \\
        HyperAtten  & \textbf{0.647 $\pm$ 0.003} & \textbf{0.670 $\pm$ 0.003}  & 0.633 $\pm$ 0.003 & \textbf{0.632 $\pm$ 0.003} & \textbf{0.661 $\pm$ 0.004} & 0.636 $\pm$ 0.004 \\
        GraphSAINT & 0.630 $\pm$ 0.000 & 0.663 $\pm$ 0.000 & 0.644 $\pm$ 0.000  & 0.625 $\pm$ 0.000 & 0.650 $\pm$ 0.000 & 0.643 $\pm$ 0.000 \\
        \bottomrule
    \end{tabular}}
\end{table}

\begin{table}[htbp]
  \caption{Accuracies of the selected GNNs on the GRAND-Diseases datasets. GraphSAGE achieves the best performance on these datasets. Hypergraph GNNs and simple graph GNNs exhibit similar performance on these datasets.}
  \label{table:grand-disease}
  \centering
  \begin{tabular}{lllll}
    \toprule
    Method      & Leukemia & LungCancer & StomachCancer & KidneyCancer \\
    \midrule
    RandomGuess & 0.333 & 0.333 & 0.333  &0.333\\
    \midrule
    GCN         & 0.582 $\pm$ 0.001 & 0.596 $\pm$ 0.001 & 0.602 $\pm$ 0.007 & 0.581 $\pm$ 0.002 \\ 
    GraphSAGE   & \textbf{0.604 $\pm$ 0.016} & \textbf{0.615 $\pm$ 0.015} & 0.602 $\pm$ 0.016 & \textbf{0.596 $\pm$ 0.014} \\
    GAT         & 0.587 $\pm$ 0.005 & 0.596 $\pm$ 0.000 & 0.600 $\pm$ 0.007 & 0.581 $\pm$ 0.003 \\
    GATv2       & 0.583 $\pm$ 0.000 & 0.591 $\pm$ 0.005 & 0.596 $\pm$ 0.002 & 0.579 $\pm$ 0.006 \\
    HyperConv   & 0.586 $\pm$ 0.003 & 0.593 $\pm$ 0.003 & 0.596 $\pm$ 0.004 & 0.577 $\pm$ 0.006 \\
    HyperAtten  & 0.593 $\pm$ 0.008 & 0.608 $\pm$ 0.008 & \textbf{0.604 $\pm$ 0.022} & 0.578 $\pm$ 0.012 \\
    GraphSAINT  & 0.583 $\pm$ 0.000 & 0.595 $\pm$ 0.000 & 0.596 $\pm$ 0.000 & 0.582 $\pm$ 0.000 \\
    \bottomrule
  \end{tabular}
\end{table}

\begin{table}[htbp]
  \caption{Accuracies of the selected GNNs the Amazon datasets. GraphSAINT performs the best among all GNNs on these datasets, and hypergraph GNNs tend to outperform simple graph GNNs.}
  \label{table:amazon}
  \centering
  \begin{tabular}{lll}
    \toprule
    Method      & Computers & Photos\\
    \midrule
    RandomGuess & 0.100 & 0.100 \\
    \midrule
    GCN         & 0.756 $\pm$ 0.041 & 0.295 $\pm$ 0.017 \\
    GraphSAGE   & 0.582 $\pm$ 0.108 & 0.366 $\pm$ 0.061 \\
    GAT         & 0.742 $\pm$ 0.043 & 0.434 $\pm$ 0.074 \\
    GATv2       & 0.566 $\pm$ 0.046 & 0.420 $\pm$ 0.075 \\
    HyperConv   & 0.842 $\pm$ 0.020 & 0.337 $\pm$ 0.059 \\
    HyperAtten  & 0.663 $\pm$ 0.005 & 0.465 $\pm$ 0.033 \\
    GraphSAINT  & \textbf{0.875 $\pm$ 0.020} & \textbf{0.512 $\pm$ 0.141} \\
    \bottomrule
  \end{tabular}
\end{table}

\begin{table}[htbp]
  \caption{MSEs ($\downarrow$) of the selected GNNs the MUSAE-Wiki datasets. GraphSAINT performs the best on these datasets, and hypergraph GNNs tend to perform at the same level as simple graph GNNs.}
  \label{table:wiki}
  \centering
  \begin{tabular}{llll}
    \toprule
    Method     & Chameleon & Squirrel & Crocodile \\
    \midrule
    GCN        & 7.319 $\pm$ 0.000 & 8.761 $\pm$ 0.001 & 6.779 $\pm$ 0.005 \\
    GraphSAGE  & 6.945 $\pm$ 0.005 & 8.310 $\pm$ 0.003 & 6.380 $\pm$ 0.005 \\
    GAT        & 6.557 $\pm$ 0.154 & 8.093 $\pm$ 0.054 & 6.249 $\pm$ 0.261 \\
    GATv2      & 7.290 $\pm$ 0.019 & 8.600 $\pm$ 0.011 & 6.717 $\pm$ 0.005 \\
    HyperConv  & 7.230 $\pm$ 0.002 & 8.706 $\pm$ 0.000 & 6.712 $\pm$ 0.001 \\
    HyperAtten & 7.451 $\pm$ 0.000 & 8.782 $\pm$ 0.000 & 6.942 $\pm$ 0.000 \\
    GraphSAINT & \textbf{5.165 $\pm$ 0.027} & \textbf{7.541 $\pm$ 0.023} & \textbf{4.898 $\pm$ 0.035} \\
    \bottomrule
  \end{tabular}
\end{table}

\begin{table}[htbp]
    \caption{Statistics of the subgraphs obtained by various samplers. HybridGraphSAINT-Edge and HybridGraphSAINT-RW perform the best in preserving the original hybrid graph characteristics.}
    \label{table:subgraphStats}
    \centering
    \resizebox{\textwidth}{!}{\begin{tabular}{lrrrrrr}
        \toprule
        &\multicolumn{3}{c}{\makecell{MUSAE-GitHub}}&\multicolumn{3}{c}{\makecell{MUSAE-Facebook}}\\
        \midrule
        Samplers & \makecell[r]{Avg.\\Node\\Degree} & \makecell[r]{Avg.\\Hyperedge\\Degree} & \makecell[r]{Clustering\\Coef.}& \makecell[r]{Avg.\\Node\\Degree} & \makecell[r]{Avg.\\Hyperedge\\Degree} & \makecell[r]{Clustering\\Coef.} \\
        \midrule
        Original & \textit{30.66} & \textit{0.168} & \textit{4.590} &  \textit{30.44} & \textit{0.360} & \textit{9.905} \\
        \midrule
        HybridGraphSAINT-Node & 42.59 & 0.268 & 13.20 & 42.72 & 0.268 & 13.11   \\
        HybridGraphSAINT-Edge & 38.63 & 0.366 & 11.42 & 38.66 & \textbf{0.366} & \textbf{11.50} \\
        HybridGraphSAINT-RW   & \textbf{35.35} & \textbf{0.203} & \textbf{6.48} &  \textbf{35.03} & 0.206 & 6.54 \\
        RandomNode            &  2.25 & 0.020 & 19.16 &  2.58 & 0.037 & 21.73 \\
        RandomHyperedge       & 72.38 & 0.282 &  7.48 & 72.74 & 0.279 &  7.44 \\
        \bottomrule     
    \end{tabular}}
\end{table}
\section{Data Accessibility}

The source code and full datasets of HGB is publicly available at \url{https://zehui127.github.io/hybrid-graph-benchmark/}. While the raw dataset in JSON format is hosted at \url{https://zenodo.org/record/7982540}, we recommend the users to access the datasets through our Python library \texttt{hybrid-graph}, which is installable via \texttt{pip}. This would allow the users to read the hybrid graphs in the format of PyTorch Geometric \texttt{Data} objects.

\clearpage
\section{Licence} \label{appendix:licence}

The raw data for the MUSAE datasets are licenced under the the GNU General Public Licence, version 3 (GPLv3)\footnote{\label{gpl}\url{https://www.gnu.org/licenses/gpl-3.0.html}}. The raw data for the GRAND datasets are licenced under the Creative Commons Attribution-ShareAlike 4.0 International Public Licence (CC BY-SA 4.0)\footnote{\url{https://creativecommons.org/licenses/by-sa/4.0/}}. The raw data for the Amazon datasets are licenced under the Amazon Service licence\footnote{\label{amazon-licence}\url{https://s3.amazonaws.com/amazon-reviews-pds/LICENSE.txt}}. Having carefully observed the licence requirements of all data sources and code dependencies, we apply the following licence to our source code and datasets:

\begin{itemize}
    \item The source code of HGB is licenced under the MIT licence\footnote{\url{https://opensource.org/license/mit/}};
    \item The MUSAE and GRAND datasets are licenced under the GPLv3 licence\textsuperscript{\ref{gpl}};
    \item The Amazon datasets are licenced under the Amazon Service licence\textsuperscript{\ref{amazon-licence}}.
\end{itemize}

\section{Ethics Statement}

All datasets constructed in HGB are generated from public open-source datasets, and the original raw data downloaded from the data sources do not contain any personally identifiable information or other sensitive contents. The node prediction tasks for the HGB datasets are designed to ensure that they do not, by any means, lead to discriminations against any social groups. Therefore, we are not aware of any social or ethical concern of HGB. Since HGB is a general benchmarking tool for representation learning on complex graphs, we also do not forsee any direct application of HGB to malicious purposes. However, the users of HGB should be aware of any potential negative social and ethical impacts that may arise from their chosen downstream datasets or tasks outside of HGB, if they intend to use the HGB datasets as pre-training datasets to perform trasnfer learning.

\end{document}